\documentclass[12pt]{article}
\usepackage{amsmath}
\usepackage{amsthm}
\usepackage{amssymb}
\usepackage{mathtools}

\newcommand{\bb}{\mathbb}
\renewcommand{\Box}{\qedhere}

\renewcommand{\glossary}[2]{}
\def\addcontentsline#1#2#3{} 

\newtheorem{thm}{Theorem}

\newtheorem{lem}[thm]{Lemma}

\newtheorem{defn}{Definition}

\newcommand{\ie}{{\em i.e.\ }}

\newcommand{\eg}{{\em e.g.\ }}

\newcommand{\Shat}{{\hat S}}

\newcommand{\fhat}{{\hat f}}

\newcommand{\be}{\begin{equation*}}

\newcommand{\ep}{{\varepsilon}}

\newcommand{\N}{{\cal N}}
\newcommand{\Nco}{{\N_\co}}

\newcommand{\T}{{\bb T}}

\renewcommand{\(}{\left(}
\renewcommand{\)}{\right)}
\renewcommand{\[}{\left[}
\renewcommand{\]}{\right]}

\renewcommand{\hbar}{{\overline h}}

\newcommand{\co}{{\text{\rm co}}}

\newcommand{\lin}{{\text{\rm lin}}}
\newcommand{\sign}{{\text{\rm sign}}}

\newcommand{\cobar}[1]{{\overline{\text{\rm co}\(#1\)}}}
\newcommand{\linbar}[1]{{\overline{\text{\rm lin}\(#1\)}}}

\newcommand{\R}{{\mathbb R}}

\begin{document}
\title{A result relating convex n-widths to covering numbers with some 
applications to neural networks}
\author{Jonathan Baxter and Peter Bartlett\\ 
        Department of Systems Engineering \\ 
Australian National University \\ Canberra 0200, Australia}
\date{}
\maketitle

\abstract{In general, approximating classes of functions defined over
high-dimensional input spaces by linear combinations of a fixed set of basis
functions or ``features'' is known to be hard. Typically, the worst-case 
error of the best basis set decays only as fast as $\Theta\(n^{-1/d}\)$,
where $n$ is the number of basis functions and $d$ is the input dimension. 
However, there are many examples of high-dimensional pattern recognition 
problems (such as face recognition) where linear combinations of small sets 
of features do solve the problem well. Hence these function classes
do not suffer from the ``curse of dimensionality'' associated with more
general classes. It is natural then, to look for characterizations of
high-dimensional function classes that nevertheless are approximated well by 
linear combinations of small sets of features. In this
paper we give a general result relating the error of approximation of a
function class to the covering number of its ``convex core''.
For one-hidden-layer neural networks, covering numbers of the class of
functions computed by a single hidden node upper bound the covering numbers
of the convex core. Hence, using standard results we obtain upper bounds on
the approximation rate of neural network classes.}

\section{Introduction}
A common approach to solving high-dimensional pattern recognition problems
is to choose a small set of relevant features, linear combinations of which
approximate well the functions in the target class. 
Another way of viewing the small set of features is as a low-dimensional
representation of the input space. Recently there have been a number of papers
showing how such feature sets can be {\em learnt} 
(\eg \cite{JB95a,Thrun94,Edelman96}) 
using neural network and related techniques. 

The assumption that a class of functions can be approximated well by linear 
combinations of a small set of features is a very strong one, for it is
known that approximating the class 
of functions with domain
$[0,1]^d$ and bounded first derivative to within error $\ep$ requires at
least $1/\ep^d$ basis functions. 
The purpose of this paper is to gain a better understanding of the factors 
governing 
whether a function class possesses a low-dimensional representation or not.
A theorem is presented showing that the dimension of the representation
required to achieve an approximation error of less than $\ep$ is closely
related to the size of the smallest $\ep$-covering number of the ``convex core''
of the function class. We show that the covering numbers of the 
convex core of a single-hidden-layer neural
network are upper-bounded by covering numbers of the 
class of functions computed by a single hidden node, which enables
an easy calculation of upper bounds on the 
approximation error in this case. We give upper bounds for several hidden
node classes, including general VC classes, linear threshold classes
and classes smoothly paramaterized by several real variables.

The remainder of the paper is organised as follows. Introductory definitions are 
presented in section \ref{defsec} along with some simple results. The main
theorem is given in section \ref{thmsec}, along with examples showing
that it is tight. Applications of the theorem to neural networks are given
in section \ref{appsec}. 

\section{Preliminary definitions}
\label{defsec}
Let $\(X, \|\cdot\|\)$ be a Banach space.
For any $S\subset X$ and any $f\in X$, define 
$$
\|f - S\|:= \inf_{s\in S} \|f - s\|.
$$
$S$ is {\em convex} if 
$s_1,\dots,s_n \in S$ implies $\sum_{i=1}^n \lambda_i s_i \in S$ 
for any $\lambda_1,\dots,\lambda_n$ such that 
$\sum_{i=1}^n |\lambda_i| \leq 1$.
If $S\subseteq X$, define 
$$
\co_n(S):= \left\{\sum_{i=1}^n \lambda_i s_i:
\sum_{i=1}^n|\lambda_i| \leq 1, s_i\in S, i=1,\dots,n\right\},
$$
\ie $\co_n(S)$ is
the set of all convex combinations of $n$ elements from $S$. Let 
$\cobar{S}$ denote the set of
all $f\in X$ such that $\lim_{n\rightarrow\infty} \|f - \co_n(S)\| = 0$.
Similarly, define 
$$
\lin_n(S):= \left\{\sum_{i=1}^n \lambda_i s_i:
s_i\in S, i=1,\dots,n\right\},
$$
and let $\linbar{S}$ denote the set of all $f\in X$ such that 
$\lim_{n\rightarrow\infty} \|f - \lin_n(S)\| = 0$.

\begin{defn}
\label{convexwidth}
For any $K\subseteq X$, define the {\em convex n-width} of $K$ by
$$
c_n(K) := \inf_{\phi_1,\dots,\phi_n\in X}\sup_{f\in K} \left\|f - 
\co_n(\{\phi_1,\dots,\phi_n\})\right\|.
$$
\end{defn}
Note that $c_n(K)$ is similar to the {\em Kolmogorov n-width} $d_n(K)$ 
\cite{Pinkus}, except that instead of arbitrary 
linear combinations of the $\phi_1,\dots,\phi_n$, the approximation 
has to be achieved with convex combinations. 
\begin{defn}
\label{nwidth}
The {\em Kolmogorov n-width} of $K\subseteq X$ is 
$$
d_n(K) := \inf_{\phi_1,\dots,\phi_n\in X}\sup_{f\in K} \left\|f - 
\lin_n(\{\phi_1,\dots,\phi_n\})\right\|.
$$
\end{defn}

Let $\N(\ep,K)$ denote the size of the 
smallest $\ep$-cover of $K$, \ie the size of the smallest set 
$\left\{\phi_1,\dots,\phi_N\right\} \subset X$
such that for all $f\in K$ there exists $\phi_i$ such
that $\|f - \phi_i\| \leq \ep$. If there is no such set then set 
$\N(\ep,K) := \infty$.
\begin{defn}
\label{core}
For any $K\subseteq X$, define
\begin{equation}
\label{nco}
\Nco(\ep,K) := \min_{S\subseteq X: \cobar{S} \supseteq K} \N(\ep, S).
\end{equation}
If $S\subseteq X$ is such that $\cobar{S} \supseteq K$ and 
$\N(\ep, S) = \Nco(\ep,K)$, then we say $S$ is a {\em convex $\ep$-core} 
of $K$. 
\end{defn}
Note that 
there always exists at least one
convex $\ep$-core of $K$ (if $\Nco(\ep,K) = \infty$ then we can set 
$S = K$, otherwise the fact that $\N(\ep,S)$ is integer 
valued means that the $\min$ in  \eqref{nco} must be attained for some $S$).

\section{Results}
\label{thmsec}
\begin{thm}
\label{corethm}
For any set $K\subseteq X$ and for all $\ep >0$,
\begin{enumerate}
\item \label{part1}
if $n < \Nco(\ep, K) - 1$ then $c_n(K) > \ep$,
\item \label{part2}
if $n \geq \Nco(\ep,K)$ then $c_n(K) \leq \ep$.
\end{enumerate}
\end{thm}
{\bf Proof}

\noindent{\em Part \ref{part1}:} We show $c_n(K) \leq \ep$ implies 
$n + 1 \geq \Nco(\ep,K)$. Fix $\delta > 0$. If $c_n(K) \leq \ep$ then
there exists $\phi_1,\dots,\phi_n$ such that for all $f\in K$ there exists
$\lambda_1,\dots,\lambda_n$ with $\sum_{i=1}^n |\lambda_i| \leq 1$ and 
$$
\left\|f - \sum_{i=1}^n \lambda_i \phi_i\right\| \leq \ep + \delta.
$$
Set $\phi_0 := 0$, $\alpha:= \ep +\delta$ and let 
$$
S := \bigcup\limits_{i=0}^n B_\alpha\(\phi_i\),
$$
where $B_\alpha(\phi_i) := \{f \in X\colon \|f - \phi_i\| \leq \alpha\}$.
Note that $\N(\alpha,S) \leq n + 1$. We show that $\cobar{S} \supseteq K$ 
and so $\Nco(\alpha, K) \leq \N(\alpha,S) \leq n+1$. Letting 
$\delta \rightarrow 0$ (\ie $\alpha \rightarrow \ep^+)$ gives the result 
because the covering numbers are continuous from the right.

So fix $f\in K$ and choose a convex combination $\sum_{i=1}^n\lambda_i \phi_i$
so that $\|f - \sum_{i=1}^n \lambda_i \phi_i\| \leq \alpha$ as above.
Set 
$\lambda_0 := 1 - \sum_{j=1}^n |\lambda_j|$ and 
for each $i=0,\dots,n$,
$$
\phi'_i := \phi_i + \sign(\lambda_i)\(f - \sum_{j=1}^n \lambda_j\phi_j\)
$$
Note that $\sum_{j=0}^n |\lambda_j| = 1$ and for each $i=0,\dots,n$, 
$$
\left\|\phi_i - \phi'_i \right\| = 
\left\|f - \sum_{j=1}^n \lambda_j \phi_j \right\| \leq \alpha,
$$
hence $\phi'_i \in S$ for each $i=0,\dots,n$. Now,
\begin{align*}
\left\| f - \sum_{i=0}^n \lambda_i \phi'_i \right\| &= 
\left\| f - \lambda_0\(f - \sum_{j=1}^n\lambda_j \phi_j\) - \right.\\
&\qquad\left.\sum_{i=1}^n \lambda_i\[\phi_i + \sign(\lambda_i)
\(f - \sum_{j=1}^n \lambda_j\phi_j\)\]\right\| \\
&= \left\| \(1 - \lambda_0 - \sum_{i=1}^n |\lambda_i |\)f + 
\sum_{i=1}^n \lambda_i\phi_i \(\lambda_0 - 1 + 
\sum_{j=1}^n |\lambda_j|\)\right\|\\
& = 0,
\end{align*}
and so $f\in \cobar{S}$, as required
(we have actually proved the stronger result 
that $f\in\co_{n+1}(S)$, but I'm not sure if that helps). \\

\noindent{\em Part \ref{part2}:} We show that
if $n = \Nco(\ep, K)$ then $c_n(K) \leq \ep$.  Part \ref{part2} then
follows because if $n' > n$ then $c_{n'}(K) \leq c_n(K)$. So let $n =
\Nco(\ep,K)$ and let $S$ be a convex $\ep$-core of $K$ (recall by the
remarks following definition \ref{core} that such an $S$ always
exists). By definition, there exists $\Shat :=
\{\phi_1,\dots,\phi_n\}\subset X$ such that for all $s\in S$ there
exists $\phi_i\in \Shat$ such that $\| s - \phi_i\| \leq \ep$. As
$\cobar{S} \supseteq K$, for all $f\in K$ and $\delta > 0$ we can find
$f_1,\dots,f_N \in S$ and $\lambda_1,\dots,\lambda_N$ such that
$\sum_{i=1}^N |\lambda_i| \leq 1$ and $\|f - \sum_{i=1}^N \lambda_i
f_i\| \leq \delta$. For each $i=1,\dots, N$, choose $\fhat_i\in \Shat$
such that $\|f_i - \fhat_i\| \leq \ep$. Note that each $\fhat_j$ 
is one of the $\phi_i$'s, so for each $i=1,\dots,n$ 
define
$\lambda'_i := \sum_{j : \fhat_j = \phi_i}\lambda_j$.
Note that $\sum_{i=1}^n |\lambda'_i| \leq \sum_{i=1}^N |\lambda_i| \leq 1$, 
hence $\sum_{i=1}^n \lambda'_i \phi_i$ is a convex combination of the 
$\phi_i$'s. Now,
\begin{align*}
\left\|f - \sum_{i=1}^n\lambda'_i \phi_i\right\| &= 
\left\|f - \sum_{i=1}^N \lambda_i \fhat_i\right\| \\
&\leq \left\|f - \sum_{i=1}^N\lambda_i f_i \right\| + \left\|\sum_{i=1}^N \lambda_i f_i 
- \sum_{i=1}^N \lambda_i \fhat_i \right\| \\
&\leq \delta + \sum_{i=1}^N |\lambda_i | \left\|f_i - \fhat_i\right\| \\
&\leq \delta + \ep.
\end{align*}
Letting $\delta \rightarrow 0$ shows that $c_n(K) \leq \ep$, as required.
$\Box$

\subsection{Theorem \ref{core} is as tight as possible}
There is a gap of ``1'' in theorem \ref{core}, in the sense that
when $n = \Nco(\ep,K) - 1$ we don't
know whether $c_n(K) \leq \ep$ or $c_n(K) > \ep$. 
In this section we show the gap is necessary: there exist classes for 
which $c_{\Nco(\ep,K) - 1}(K) > \ep$ and also classes for which 
$c_{\Nco(\ep,K) - 1}(K) \leq \ep$. The first case is trivial; just consider 
the class consisting of a single non-zero element $K = \{\phi \neq 0\}$. 
Clearly $\Nco(\ep,K) = 1$ for all $\ep \geq 0$, but $c_0(K) = \|\phi\| > 0$.

To demonstrate the second case we need a lemma.
\begin{lem}
\label{rellem}
If $c_n(K) < \infty$ then $c_n(K) = d_n(K)$.
\end{lem}
\noindent{\bf Proof}\\
For any $S\subseteq X$, define
$$
\lin_n^\Lambda(S) := \left\{\sum_{i=1}^n\lambda_i s_i\colon \sum_{i=1}^n 
|\lambda_i| \leq \Lambda, s_i \in S, i=1,\dots,n\right\}.
$$
Let $\ep = d_n(K) \leq c_n(K) < \infty$ 
and suppose that for all $\delta > 0$ there exists 
$\phi_1,\dots,\phi_n\in X$ and $\Lambda < \infty$ such that 
$$
\sup_{f\in K} \left\| f - \lin_n^\Lambda(\{\phi_1,\dots,\phi_n\})\right\|
< \ep + \delta.
$$
Then $c_n(K) = \ep$ also because $\co_n(\Lambda S) = \lin_n^\Lambda(S)$.
So we need only consider the case in which there exists $\delta > 0$ 
such that for all $\phi_1,\dots,\phi_n$ and $\Lambda < \infty$,
$$
\sup_{f\in K} \left\| f - \lin_n^\Lambda(\{\phi_1,\dots,\phi_n\})\right\|
\geq \ep + \delta.
$$
Equivalently, we assume that for all 
$S = \{\phi_1,\dots,\phi_n\}$ for which  $\sup_{f\in k} \|f - \lin_n(S)\| < 
\ep + \delta$, there exists a sequence $(f_N)_{N=1}^\infty$ in $K$  such that 
if  $\lambda_{N1},\dots,\lambda_{Nn}$  satisfy
\begin{equation}
\label{blah}
\left\| f_N - \sum_{i=1}^n \lambda_{Ni} \phi_i \right\| < \ep + \delta
\end{equation}
then $\sum_{i=1}^n |\lambda_{Ni}| > N$. 
Fix such an $S = \{\phi_1,\dots,\phi_n\}$, a sequence $(f_N)_{N=1}^\infty$,
and for each $f_N$ a set of coefficients 
$\lambda_{N1},\dots,\lambda_{Nn}$  satisfying \eqref{blah}.
We show that if $c_n(K) < \infty$ this leads to a contradiction.
Note that without loss of generality we can assume $\phi_1,\dots,\phi_n$ 
are linearly independent.

As $c_n(K)  < \infty$, there exists $\psi_1,\dots,\psi_n$ such that 
$$
\sup_{f\in K}\left\|f - \co_n(\{\psi_1,\dots,\psi_n\})\right\| 
= \beta,
$$
for some $\beta < \infty$. So for each $f_N$, 
choose $\lambda'_{N1},\dots,\lambda'_{Nn}$  such that 
$\sum_{i=1}^n |\lambda'_{Ni}| \leq 1$ and 
\begin{equation}
\label{b1}
\left\| f_N - \sum_{i=1}^n \lambda'_{Ni} \psi_i \right\| \leq \beta.
\end{equation}
Equations \eqref{blah} and \eqref{b1} imply that 
\begin{equation}
\label{b2}
\left\|\sum_{i=1}^n \lambda_{Ni} \phi_i - 
\sum_{i=1}^n \lambda'_{Ni} \psi_i\right\| < \beta + \ep + \delta.
\end{equation}
In addition,
\begin{align*}
\left\|\sum_{i=1}^n \lambda_{Ni} \phi_i - 
\sum_{i=1}^n \lambda'_{Ni} \psi_i\right\| &\geq
\left\|\sum_{i=1}^n \lambda_{Ni} \phi_i\right\| - 
\left\|\sum_{i=1}^n \lambda'_{Ni} \psi_i\right\| \\
&\geq 
\left\|\sum_{i=1}^n \lambda_{Ni} \phi_i\right\| - 
\left\|\psi_{\text{\rm max}}\right\|,
\end{align*}
where $\left\|\psi_{\text{\rm max}}\right\| := \max_i 
\left\|\psi_i\right\|$. However, 
$\left\|\sum_{i=1}^n \lambda_{Ni} \phi_i\right\| \rightarrow \infty$
as $N\rightarrow\infty$\footnote{To see this set 
$$
a:= \inf_{\alpha_1,\dots,\alpha_n:\sum|\alpha_i| = 1}
\left\|\sum_{i=1}^n\alpha_i \phi_i\right\|.
$$
$a > 0$ because the $\phi_1,\dots,\phi_n$ are linearly independent. 
As $\sum_{i=1}^n  |\lambda_{Ni}| > N$ (by assumption), 
$\|\sum_{i=1}^n \lambda_{Ni} \phi_i\| > Na$.}, which contradicts
\eqref{b2}. $\Box$

Now we can exhibit a class for which 
$c_{\Nco(\ep,K) - 1}(K) = c_{\Nco(\ep,K)}(K) = \ep$.
Let $\T$ denote the unit circle. The {\em Sobolev space} $W_2^r(\T)$ is
the set of all functions $f$ on $\T$ 
for which  $f^{(r-1)}$ (the $r-1$th derivative
of $f$) is absolutely continuous and 
$\|f^{(r)}\|_2 := (\int_\T \[f^{(r)}(t)\]^2 \,dt)^{1/2} < \infty$.
Define
$$
B^r_2(\T,C) := \left\{f\in W^r_2(\T)\colon \|f^{(r)}\|_2 \leq 1,
\left|\int_\T f(t) \, dt\right| \leq C\right\}.
$$
Consider approximation within the Banach space of all measurable functions
on $\T$ with norm $\|\cdot\|_2$. The following theorem 
follows easily from a similar result of Kolmogorov 
\cite{Kolmogorov36,Lorentz96}.
\begin{thm}
\label{kthm}
For $n,r=1,2,\dots$ and $C > 2\pi$,
\begin{equation}
\label{keqn}
d_{2n-1}(B^r_2(\T,C)) = d_{2n}(B^r_2(\T,C)) = n^{-r}.
\end{equation}
The subspace $\left\{1, \sin t, \cos t, \dots, \sin (n-1)t, \cos (n-1)t\right\}$
is an optimal $2n - 1$ dimensional subspace. 
\end{thm}
\begin{lem}
\label{mylem}
For all $n,r=1,2, \dots$ and $C > 2\pi$,
$$
c_n\(B^r_2(\T,C)\) = d_n\(B^r_2(\T,C)\).
$$
\end{lem}
\noindent{\bf Proof.} If $f\in B^r_2(\T,C)$ it has a representation
$$
f(t) = a_0 + \sum_{k=1}^\infty(a_k \cos kt + b_k \sin kt)
$$
with $|a_0| \leq C/(2\pi)$ and 
\begin{equation}
\label{s}
\|f^{(r)}\|_2^2 = \pi \sum_{k=1}^\infty k^{2r}(a_k^2 + b_k^2) \leq 1.
\end{equation}
$\sum_{k=1}^\infty |a_k| + |b_k|$ will be maximised subject to the constraint
\eqref{s} if $r=1$. In that case, 
a variational argument shows that if 
$$
a_k = b_k = \frac{\sqrt{3}}{\pi k^2} 
$$
then $\sum_{k=1}^\infty |a_k| + |b_k|$ is maximal, and has value
$\pi/\sqrt{3}$. Hence each optimal linear approximation will have the sum
of the absolute values of its coefficients less than $C/(2\pi) +
\pi/\sqrt{3}$, which implies that $c_n(B^r_2(\T,C)) < \infty$. Applying lemma
\ref{rellem} gives the result. $\Box$

As $c_{2n-1}(B^r_2(\T,C) = 
c_{2n}(B^r_2(\T,C)) = n^{-r}$, we know that the gap 
in theorem \ref{corethm} must be necessary, for if there was no gap the 
approximation error would have to decrease {\em every} time the number of
basis functions is increased, not every second time as is the case with this
example.

\section{Neural network applications}
\label{appsec}
In this section theorem \ref{corethm} is used to calculate upper bounds on
the convex n-widths of various function classes computed by one-hidden-layer
``neural networks'' for which the sum of the absolute values of the output 
weights is bounded by $1$. With this restriction on the output weights,
neural network classes $K_{nn}$ are equal to $\co(S)$ where $S$ is the set of all
functions computed by a single hidden node. Hence 
$\Nco(\ep,K_{nn}) \leq \N(\ep,S)$, which via theorem \ref{corethm} gives
$$
c_{\N(\ep,S)}(K_{nn}) \leq \ep.
$$
It would be nice to show that the node classes are convex $\ep$-cores, \ie
that $\Nco(\ep,K_{nn}) = \N(\ep,S)$, for then we would obtain almost tight
lower bounds on the approximate rates also. 
\begin{lem}
If $K = \cobar{S}$  then $S$ is a convex $\ep$-core of $K$ 
(\ie $\N(\ep,S) = \Nco(\ep,K)$) if and only if $\N(\ep,S) = \Nco(\ep,S)$. 
\end{lem}
\noindent {\bf Proof.}
Suppose $K=\cobar{S}$ and $S$ is a convex $\ep$-core of $K$. Let $T$ be 
a convex $\ep$-core of $S$, so trivially 
$\N(\ep,T) \leq \N(\ep,S)$. Now, $K = \cobar{S} \subseteq \cobar{\cobar{T}} = 
\cobar{T}$, which implies $\N(\ep,T) \geq \N(\ep,S)$, so $\N(\ep,T) =
\N(\ep,S)$, as required. 

Suppose $K=\cobar{S}$ and $\N(\ep,S) = \Nco(\ep,S)$. Clearly 
$\N(\ep,S) \geq \Nco(\ep,K)$. Let $T$ be a convex 
$\ep$-core of $K$, which implies $\cobar{T} \supseteq S$ and so 
$\Nco(\ep,K) = \N(\ep,T) \geq \Nco(\ep,S) = \N(\ep,S)$. $\Box$

So one way 
 to verify that the node function class $S$ is a convex $\ep$-core is to 
show that $\N(\ep,S) = \Nco(\ep,S)$, however this is more difficult than it 
looks, so we present only the upper bounds.

\subsection{VC classes S}
Let $\N\(\ep,S,L_p(P)\)$ denote the smallest $\ep$-cover of $S$ under the 
$L_p(P)$ norm for some $1\leq p < \infty$ and some distribution $P$.
Let $V(S)$ denote the VC dimension of $S$.
Haussler \cite{Haussler95} proved:
$$
\N\(\ep,S,L_p(P)\) \leq K V(S) (4e)^{V(S)} \(\frac1\ep\)^{p(V(S) - 1)}
$$
(see also \cite{vanderVaart96}, theorem 2.6.4). Hence,
$$
c_n(K_{nn}(S)) \leq \frac{C}{n^{\frac1{p(V(S) - 1)}}},
$$
where $C = \(K V(S) (4e)^{V(S)}\)^{1/(p(V(S) - 1))}$. So neural networks whose node classes have
small VC dimension can be well approximated by convex combinations of fixed
sets of basis functions. 
\subsection{Linear Threshold S}
This is a special case of the previous section. The VC dimension of the
class of linear threshold functions on $\R^d$ is $d+1$ and so for the $L_p(P)$
norm 
$$
c_n(K) \leq \frac{C}{n^{\frac1{p(d+1)}}}.
$$
\subsection{Smoothly parameterized classes}
Suppose $S$ is indexed by $k$ real parameters, that is there exists
a ``mother function'' $f\colon \R^d\times [0,1]^k\to [0,1]$ such that 
$S = \{f(\cdot, y)\colon y \in [0,1]^k\}$. If we assume that the
parameterization satisfies $\|f(\cdot,y) - f(\cdot,y')\| \leq \|y - y'\|$ for
all $y,y'\in [0,1]^k$, then $\N(\ep,S) \leq (1/\ep)^k$ which gives an
approximation rate of
$$
c_n(K) \leq \frac{1}{n^{\frac1k}}.
$$

\section*{Acknowledgements}
Thanks to Mostafa Golea for helpful discussions.
\bibliographystyle{plain}
\bibliography{bib}
\end{document}